\documentclass{article}

\usepackage[nonatbib, preprint]{neurips_2024}

\usepackage[utf8]{inputenc} 
\usepackage[T1]{fontenc}    
\usepackage[pagebackref=false,breaklinks=true,letterpaper=true,colorlinks,urlcolor=blue,citecolor=blue,linkcolor=blue,bookmarks=false]{hyperref}
\usepackage{url}            
\usepackage{booktabs}       
\usepackage{amsfonts}       
\usepackage{nicefrac}       
\usepackage{microtype}      
\usepackage{xcolor}         

\usepackage{multirow}
\usepackage{multicol}
\usepackage{wrapfig}
\usepackage{mathtools}
\usepackage{enumitem}
\usepackage{arydshln}
\usepackage{bm}

\include{preamble}

\title{Expressive Gaussian Human Avatars\\ from Monocular RGB Video}

\author{%
  \textbf{Hezhen Hu}$^{1}$ \quad
  \textbf{Zhiwen Fan}$^{1}$ \quad
  \textbf{Tianhao Wu}$^{2}$ \quad
  \textbf{Yihan Xi}$^{1}$ \quad
  \textbf{Seoyoung Lee}$^{1}$ \\
  \textbf{Georgios Pavlakos}$^{1}$ \quad
  \textbf{Zhangyang Wang}$^{1}$ \\
  $^{1}$ University of Texas at Austin \quad
  $^{2}$ University of Cambridge \\
}

\begin{document}

\maketitle

\begin{abstract}
Nuanced expressiveness, particularly through fine-grained hand and facial expressions, is pivotal for enhancing the realism and vitality of digital human representations.
In this work, we focus on investigating the expressiveness of human avatars when learned from monocular RGB video; a setting that introduces new challenges in capturing and animating fine-grained details.
To this end, we introduce EVA, a drivable human model that meticulously sculpts fine details based on 3D Gaussians and SMPL-X, an expressive parametric human model.
Focused on enhancing expressiveness, our work makes three key contributions.
First, we highlight the critical importance of aligning the SMPL-X model with RGB frames for effective avatar learning.
Recognizing the limitations of current SMPL-X prediction methods for in-the-wild videos, we introduce a plug-and-play module that significantly ameliorates misalignment issues.
Second, we propose a context-aware adaptive density control strategy, which is adaptively adjusting the gradient thresholds to accommodate the varied granularity across body parts.
Last but not least, we develop a feedback mechanism that predicts per-pixel confidence to better guide the learning of 3D Gaussians.
Extensive experiments on two benchmarks demonstrate the superiority of our framework both quantitatively and qualitatively, especially on the fine-grained hand and facial details.
See the project website at \url{https://evahuman.github.io}
\end{abstract}

\section{Introduction}

High-quality digital avatar modeling exhibits a broad range of applications related to VR/AR, movie production, sign language, \emph{etc.}
For digital human representation, the nuanced capture of expressiveness is crucial for enriching its fidelity and vitality.
This is particularly evident in the detailed portrayal of hands and facial expressions, which enrich the emotional depth and interactive expression capability of humans avatars. 
In this work, we are dedicated to explicitly investigate expressiveness while building human avatars from monocular RGB video.
The studies task definition is adopting a monocular human video as input and learning an animated human avatar which endows multiple capabilities, \emph{e.g.} novel view or novel pose synthesis.

It is non-trivial to incorporate expressiveness into human avatars, particularly due to subtle and complex movements of hands and faces. 
Compared to the body, hand and face occupy smaller spatial sizes and exhibit unique characteristics. 
For instance, hand is highly-articulated, exhibiting fine-grained texture and self-occlusion among joints.
In the process of building human avatars, it is crucial to accurately capture the texture from monocular RGB video and perform animation effectively. 
A prerequisite for achieving this goal is accurate alignment between human pose information and the corresponding frame.
However, current off-the-shelf detectors often fail when capturing fine details, especially the hand region. 
Moreover, adapting the learning process of 3D Gaussians to accommodate the granularity differences among various body parts, while faithfully constructing each part, presents a significant challenge. 

Current studies~\cite{zhao2022humannerf,peng2021neural,jiang2022neuman,lei2023gart,GauHuman} mainly focus on learning human avatar on the body region and have made remarkable progress.
Early works~\cite{zhao2022humannerf,peng2021neural,jiang2022neuman} mainly utilize NeRF as an implicit representation but usually have the drawback of low training/inference speed.
Recently, more and more works are built on 3D Gaussian Splatting~\cite{kerbl20233d} for its effectiveness and efficiency, which could further speed up rendering to over 100fps.
GART~\cite{lei2023gart} utilizes a mixture of moving 3D Gaussians to approximate human geometry and appearance and enhances fine details with learnable forward skinning and latent bones.
GauHuman~\cite{GauHuman} proposes a new density control strategy, \emph{e.g.} split and clone with KL divergence and a new merge operation.
However, these methods do not consider the fine-grained details of hand and face, which cannot meet the requirement of expressiveness.

In this work, we introduce EVA, a drivable human model that meticulously sculpts expressive details using 3D Gaussians and human parametric model. 
Given a monocular RGB video, we extract the pose and mask information corresponding to each frame, which allows us to map each frame's human observation to a rest space. 
Once the avatar is constructed, it can be animated using linear blend skinning given a new pose, followed by rendering to the 2D human image.
To tackle new challenges introduced by expressiveness, we start by solving the misalignment issue between pose and RGB frame via a plug-and-play module.
By employing a fitting-based optimization, this module could produce more reliable pose predictions, facilitating a more robust foundation for following digital avatar modeling.

\begin{figure}[t!]
    \centering
    \includegraphics[width=\linewidth]{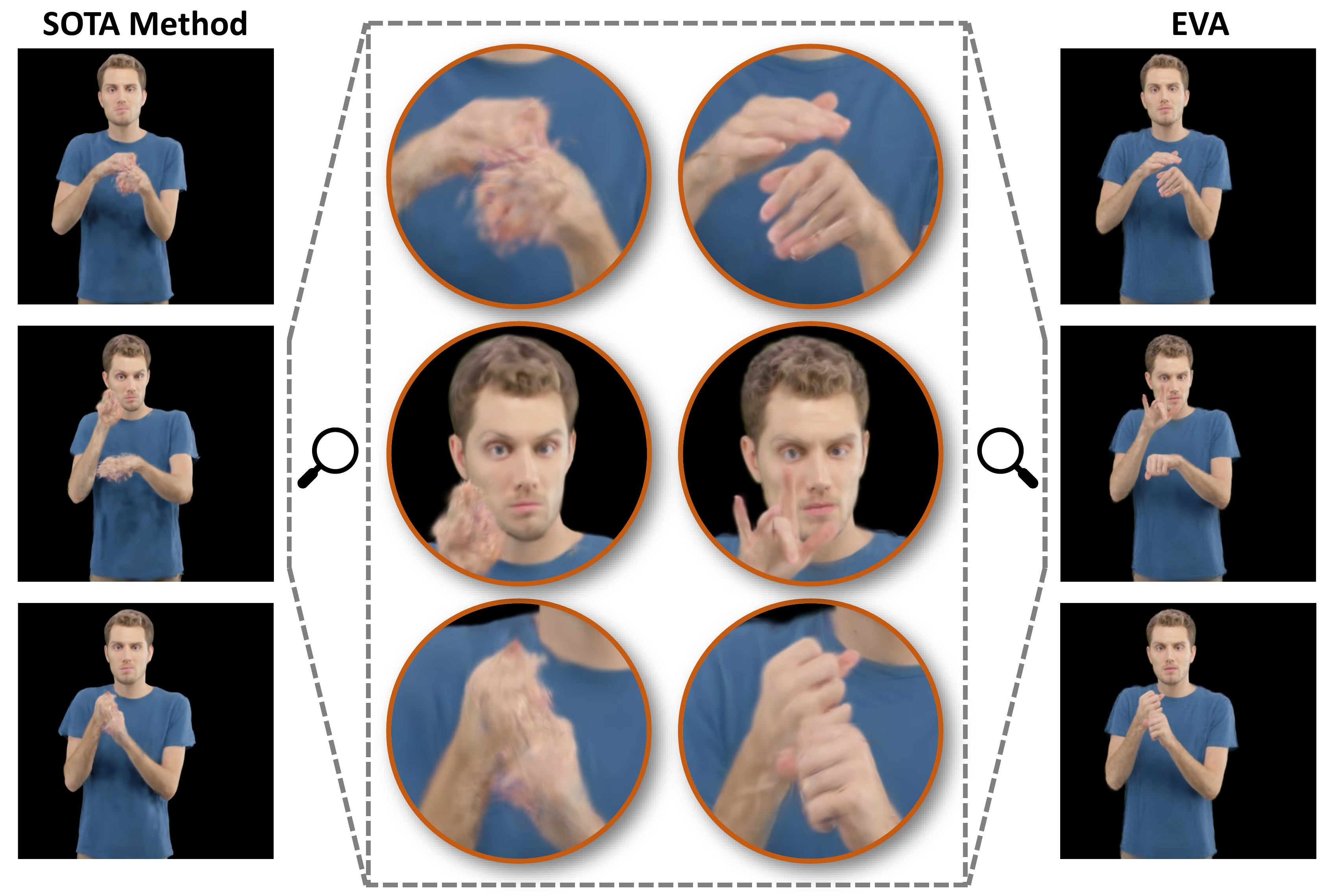}
    \caption{Qualitative comparison on novel pose synthesis between SOTA method~\cite{lei2023gart} and our EVA method.
    Given a monocular real-world video, our EVA framework generates an expressive human avatar, outperforming the previous SOTA method, especially for the hand and facial details.}
    \label{fig:intro}
     \vspace{-1.5em}
\end{figure}

Considering the granularity differences across different body parts, we propose a context-aware adaptive density control strategy for 3D Gaussian optimization. 
It leverages attributes specific to different parts and historical gradient information, to adaptively control Gaussian density. 
Furthermore, to improve the Gaussian optimization, we design a feedback mechanism, which adaptively predicts confidence scores based on the rendering result, thereby ensuring the supervision signal effectively transferring to the corresponding Gaussian. 
For evaluation on expressiveness, we build comparison baselines from related body avatar methods with a few modifications and collect a new benchmark called UPB containing in-the-wild upper body videos.
The image quality is evaluated with multiple metrics on the full, hand and face regions, respectively.
As shown in Figure~\ref{fig:intro}, we visualize the novel pose synthesis results. 
It can be observed that the rendered image by our method largely outperforms previous SOTA, with more fidelity in fine-grained details.

Our contributions are summarized as follows:
\vspace{-1.5mm}
\begin{itemize}[leftmargin=*, itemsep=0pt]
    \item In this work, we introduce EVA, an approach that can build expressive human avatars based on 3D Gaussians.
    EVA can build these avatars from monocular RGB video and through
    extensive experiments on two datasets, we demonstrate its superiority, particularly on the hand and facial details.
    \item 
    To enhance expressiveness, we propose a context-aware adaptive density control strategy to accommodate the granularity differences across human parts, followed by a feedback mechanism to better guide the 3D Gaussian optimization.
    \item To handle challenging in-the-wild videos, we propose a plug-and-play module, which critically improves the SMPL-X alignment to the RGB frames compared to off-the-shelf methods.
    We demonstrate the importance of this improvement in recovering accurate avatars.
\end{itemize}

\section{Related Work}
\subsection{Human Avatar Modeling from Monocular RGB Video}
Building the human avatar from monocular RGB video is a challenging task.
Its challenges mainly arise from dynamic capture and limited observation for each frame.
Early works~\cite{peng2021neural,Chen2023PAMI, su2021anerf, svitov2024haha, chen2021snarf, jiang2022neuman, zhao2022humannerf, yu2023monohuman,wang2022arah,xu2021h,dong2022totalselfscan,Youngjoong2023DELIFFAS,weng_humannerf_2022_cvpr} mainly resort to the combination of implicit neural representations like NeRF~\cite{mildenhall2020nerf} and parametric models to represent human avatar with high fidelity and flexibility. 
To tackle the issue of slow computation in implicit models, some works have proposed various techniques to reduce training~\cite{Geng_2023_CVPR, Jiang_2023_CVPR} or inference time~\cite{remelli2022drivable, lombardi2021mixture, chen2023primdiffusion}.
With the advent of 3D Gaussian Splatting~\cite{kerbl20233d} that achieve both high-quality and fast renderings, more and more works~\cite{lei2023gart, liu2024GVA, hu2023gaussianavatar, GauHuman, kocabas2023hugs, li2023human101, qian20233dgsavatar, jena2023splatarmor,jung2023deformable} utilize it as the base representation, in coordination with parametric human model like SMPL.
GauHuman~\cite{GauHuman} leverages human prior and KL divergence to propose a new density control strategy.
In this work, we systematically explore expressiveness while building human avatar from monocular RGB video and tackle the new challenges brought by expressiveness.

\subsection{Expressive Human Representations}
Expressiveness plays a vital role in non-verbal communication, which particularly involves hand and face regions besides the body~\cite{XAGen2023,zhang2024global,zielonka2023drivable,fu2022stylegan,fu2023unitedhuman,pavlakos2019expressive,jiang2023hifi4g,forte2023reconstructing,shen2023xavatar}.
To represent the expressive human, some parametric models are proposed, such as SMPL-X~\cite{pavlakos2019expressive}, Adam~\cite{joo2018total}, GHUM~\cite{xu2020ghum}, and \emph{etc.}
These models usually have predefined topology and can provide compact mapping from the low-dimensional embedding to the mesh representation.
However, the predefined topology limits its capability for depicting fine-grained texture.
Following works build the human avatar via neural 3D representations, jointly with the parametric model to provide human prior and animation signals.
X-Avatar~\cite{shen2023xavatar} combines NeRF and SMPL-X to build the avatar.
It further proposes part-aware sampling and initialization strategies to ensure efficient learning from high-quality 3D scan or RGB-D data.
GVA~\cite{liu2024GVA} integrates the SMPL-X model to improve the rendering image quality.
AvatarRex~\cite{zheng2023avatarrex} proposes a compositional avatar representation to separately model body, hand and faces from multi-view RGB video data.
Different from them, we aim to build the expressive human avatar with real-world monocular RGB video as input, which is a more challenging setting.

\section{Technical Approach}
In this section, we first introduce the preliminary knowledge on the SMPL-X human model and 3D Gaussian Splatting and present the general framework design on articulated human modeling.
Then we elaborate on the key technical contributions of our approach.

\subsection{Preliminaries}
\label{sec:prelim}
\noindent \textbf{SMPL-X.}
It is a parametric human body model, which extends the SMPL {body} model~\cite{loper2015smpl} with the capability of modeling both hand gestures {via MANO~\cite{MANO:SIGGRAPHASIA:2017}} and facial expressions {via FLAME~\cite{li2017flame}}.
SMPL-X can be defined as a mapping function $M(\pmb\theta, \pmb\beta, \pmb\psi): \mathbb{R}^{|\pmb\theta|} \times \mathbb{R}^{|\pmb\beta|} \times \mathbb{R}^{|\pmb\psi|} \rightarrow \mathbb{R}^{3N}$, where $\pmb\theta, \pmb\beta, \pmb\psi$ are the parameters for pose, shape, and facial expressions, respectively.
The function of SMPL-X is formulated as follows:
\begin{equation}
\label{equ:mano}
  \mathbf{M}(\pmb{\beta}, \pmb{\theta}, \pmb{\psi}) = W(\mathbf{T}(\pmb{\beta}, \pmb{\theta}, \pmb{\phi}), J(\pmb{\beta}), \pmb{\theta}, \mathbf{W}),
\end{equation}
\begin{equation}
\label{equ:mano2}
  \mathbf{T}(\pmb{\beta}, \pmb{\theta}, \pmb{\phi}) = \bar{\mathbf{T}} + B_S(\pmb{\beta}) + B_E(\pmb{\phi}) + B_P(\pmb{\theta}),
\end{equation}
where $B_P(\cdot)$, $B_S(\cdot)$ and $B_E(\cdot)$ denote pose, shape, and expression blend functions, respectively. 
$\mathbf{W}$ is a set of blend weights. 
The pose, expression and shape corrective blend shapes, \emph{i.e.,} $B_P(\pmb{\theta})$ and $B_S(\pmb{\beta})$, add corrective vertex displacements to the template human mesh $\bar{\mathbf{T}}$.
After that, linear blend skinning $W(\cdot)$~\cite{kavan2005spherical} is applied to rotate the vertices in the template mesh around the joints $J(\pmb{\beta})$ smoothed by the blend weights $\mathbf{W}$, generating the final human mesh.

\noindent \textbf{3D Gaussian Splatting.}
Previous NeRF-based methods~\cite{mildenhall2020nerf} model the scene with an implicit representation and render novel views based on volume rendering. 
In contract, 3D Gaussian Splatting~\cite{kerbl20233d}~(3DGS) models a 3D scene with a set of discrete 3D Gaussians and performs rendering through a tile-based rasterization operation, which can reach real-time rendering speeds. 
Specifically, Each Gaussian is defined with its central position $p$, 3D covariance matrix $\Sigma$, opacity $\alpha$, and color $c$ modeled by Spherical harmonics as follows:
\begin{equation}
    G(x)=exp({-\frac{1}{2}(x-p)^{T}\Sigma^{-1}(x-p)}),
\end{equation}
where $x$ is an arbitrary position in the 3D scene.
The covariance matrix $\Sigma$ is decomposed into two learnable components to make the optimization easier, $\Sigma = RSS^{T}R^{T}$, where $R$ and $S$ denote the rotation matrix and scaling vector, respectively.

During the rendering process, each 3D
Gaussian $G(x)$ is first transformed to a 2D Gaussian $G^{'}(x)$ on the image plane. 
Then a tile-based rasterizer is designed to efficiently sort the 2D Gaussians and employ $\alpha$-blending:
\begin{equation}
\begin{aligned}
    C(r)=\sum_{i\in N}c_{i}\sigma_{i}\prod_{j=1}^{i-1}(1-\sigma_{j}),~~~\sigma_{i}=\alpha_{i}G'(r),
    \label{eq:rasterization}
\end{aligned}
\end{equation}
where $r$ is the queried pixel position and N denotes the number of sorted 2D Gaussians associated with the queried pixel.
$c_{i}$ and $\alpha_{i}$ denote the color and opacity of the $i$-th Gaussian.

\noindent \textbf{Articulated 3D Human Modeling.}
We are motivated by works employing 3D Gaussian-based methods for modeling articulated objects~\cite{GauHuman,lei2023gart}. We optimize 3D Gaussians in a canonical space, which corresponds to a human in a rest pose~\cite{jiang2022neuman}. 
The Gaussians are transformed from canonical space to frame space via linear blend skinning (LBS):
\begin{align}
\mathbf{p}^f &= \mathbf{G}(\mathbf{J}^f, \bm{\theta}^f)\mathbf{p}^c + \mathbf{b}(\mathbf{J}^f, \bm{\theta}^f, \bm{\beta}^f),
\\
\bm{\Sigma}^f &= \mathbf{G}(\mathbf{J}^f, \bm{\theta}^f)\bm{\Sigma}^c + \mathbf{G}(\mathbf{J}^f, \bm{\theta}^f)^T,
\end{align}
where $\mathbf{p}^f, \bm{\Sigma}^f, \mathbf{p}^c, \bm{\Sigma}^c$ are the Gaussian mean and covariance in frame space and canonical space, respectively. 
$\mathbf{G}(\mathbf{J}^f, \bm{\theta}^f) = \sum^{K}_{k=1} w_k \mathbf{G}_k(\mathbf{J}^f, \bm{\theta}^f)$, $\mathbf{b}(\mathbf{J}^f, \bm{\theta}^f, \bm{\beta}^f) = \sum^{K}_{k=1} w_k \mathbf{b}_k(\mathbf{J}^f, \bm{\theta}^f, \bm{\beta}^f)$ are the rotation and translation, respectively, with respect to the $K$ joints, and $\mathbf{G}_k(\mathbf{J}^f, \bm{\theta}^f),  \mathbf{b}_k(\mathbf{J}^f, \bm{\theta}^f, \bm{\beta}^f)$ are the rotation and translation, respectively, with respect to the $k$-th joint. $w_k$ is the LBS weight.

To perform accurate LBS, two important components should be provided, \emph{i.e.,} LBS weight and input pose parameter. 
Learning LBS weights $w_k$ from scratch would be inefficient and can lead to a local optimum in the early stage of the training. Therefore, for each Gaussian, we start with the LBS weight of the nearest SMPL-X vertex and use an MLP $f_{\Theta_w}$ to predict an LBS weight offset $w_k^{'}$ from positionally encoded~\cite{mildenhall2020nerf} Gaussian centers $\gamma(\bm{p}^c)$. $w_k$ is therefore defined as:
\begin{align}
w_k &= \frac{
e^{\log(w_k^{\text{SMPL-X}} + \epsilon) + w_k^{'}}
}{
\sum^K_{j=1} e^{\log(w_j^{\text{SMPL-X}} + \epsilon) + w_j^{'}}
}
\\
w_k^{'} &= f_{\Theta_w}(\gamma(\bm{p}^c)[k]) 
\end{align}
\noindent
where $w_k^{\text{SMPL-X}}$ is the LBS weight of nearest SMPL-X vertex, $\epsilon = 10^{-8}$.

Starting from the input pose $\bm{\theta}^{\text{SMPL-X}}$, we further fine-tune it via a MLP-based network $f_{\Theta_{\bm{\theta}}}$, which is jointly optimized with the 3D Gaussian optimization process. 
The actual poses $\bm{\theta}$ used in optimization and rendering are therefore obtained as follows:
\begin{align}
\bm{\theta} = \bm{\theta}^{\text{SMPL-X}} \otimes f_{\Theta_{\bm{\theta}}} (\bm{\theta}^{\text{SMPL-X}})
\end{align}
\noindent
where $\otimes$ represents the vector pointwise product.

\subsection{SMPL-X Alignment for Real-World Video}
An important requirement for the recovery of accurate human avatars is to initialize the optimization with a reliable SMPL-X estimate.
However, this can be challenging, given the limitations of current off-the-shelf methods for SMPL-X fitting.
This motivates us to propose a robust fitting procedure that leverages multiple sources as pseudo Ground Truth (GT), including the initial estimation of camera parameter, SMPL-X parameters, 2D keypoints and 3D hand parameters from off-the-shelf tools~\cite{pavlakos2023reconstructing,yang2023effective,cai2024smpler}.
The SMPL-X fitting procedure minimizes the following objective:
\begin{equation}
\begin{aligned}
   \mathcal L(\pmb\theta, \pmb\beta, \pmb\phi)= \mathcal L_{2D} + \lambda_{bp}\mathcal L_{bp} + \lambda_{hp}\mathcal L_{hp}.
\label{eq:fit_loss_term}
\end{aligned}
\end{equation}

The 2D keypoint constraint term derives the supervision signal from detected 2D keypoints, which is formulated as follows:
\begin{equation}
\begin{aligned}
    \mathcal L_{2D}= \sum_{i\in J}\gamma_i\omega_i \psi(\Pi_{\pmb K}(R_{\pmb\theta}(J({\pmb\beta}))_i)-J^{2D}_{i}),
\label{eq:2d}
\end{aligned}
\end{equation}
where $\Pi_K(\cdot)$ represents the camera projection under the given camera intrinsic parameter $\pmb K$.
$\psi(\cdot)$ is the robust Geman-McClure error function which helps prevent the disturbance from noisy supervision signals.
$R_\theta(\cdot)$ denotes the function which rotates the $J(\pmb\beta)$ given the pose $\pmb\theta$.
$J^{2D}$ is the detected 2D keypoints from the method~\cite{yang2023effective}.
$\gamma$ and $\omega$ are the weighting parameters from predefined weight and detection confidence.

Our prior terms contain two components which focus on the coarse-grained body and fine-grained hand, respectively.
For the body part, we utilize the variational human body pose prior~\cite{pavlakos2019expressive} to filter infeasible body poses.
The variational pose prior can provide a compact mapping from low-dimensional embedding $\eta$ to rotation matrices of the body pose $\theta_b$.
To better optimize the low-dimensional embedding $\eta$, the estimated body pose $J^{3D}_{b}$~\cite{cai2024smpler} is treated as guidance, in coordination with the added regularization term.
The body prior loss term is formulated as follows:
\begin{equation}
\begin{aligned}
    \mathcal L_{bp}= \psi(R_\theta (J_{b}(\beta))-J^{3D}_{b})  + ||\eta||^2.
\label{eq:bp}
\end{aligned}
\end{equation}

For the hand part, we utilize the hand estimation $J^{3D}_{h}$ from the method~\cite{pavlakos2023reconstructing} to provide a better hand spatial relationship, which is formulated as follows:
\begin{equation}
\begin{aligned}
    \mathcal L_{hp}= \psi_{z}(R_\theta (J_{h}(\beta))-J^{3D}_{h}) 
\label{eq:hp}
\end{aligned}
\end{equation}
where the index $h$ represents the joints corresponding to the hand.
$\psi_{z}$ means that the robust error function only takes the z-axis coordinates into the calculation.

\begin{figure}[t!]
    \centering
    \includegraphics[width=\linewidth]{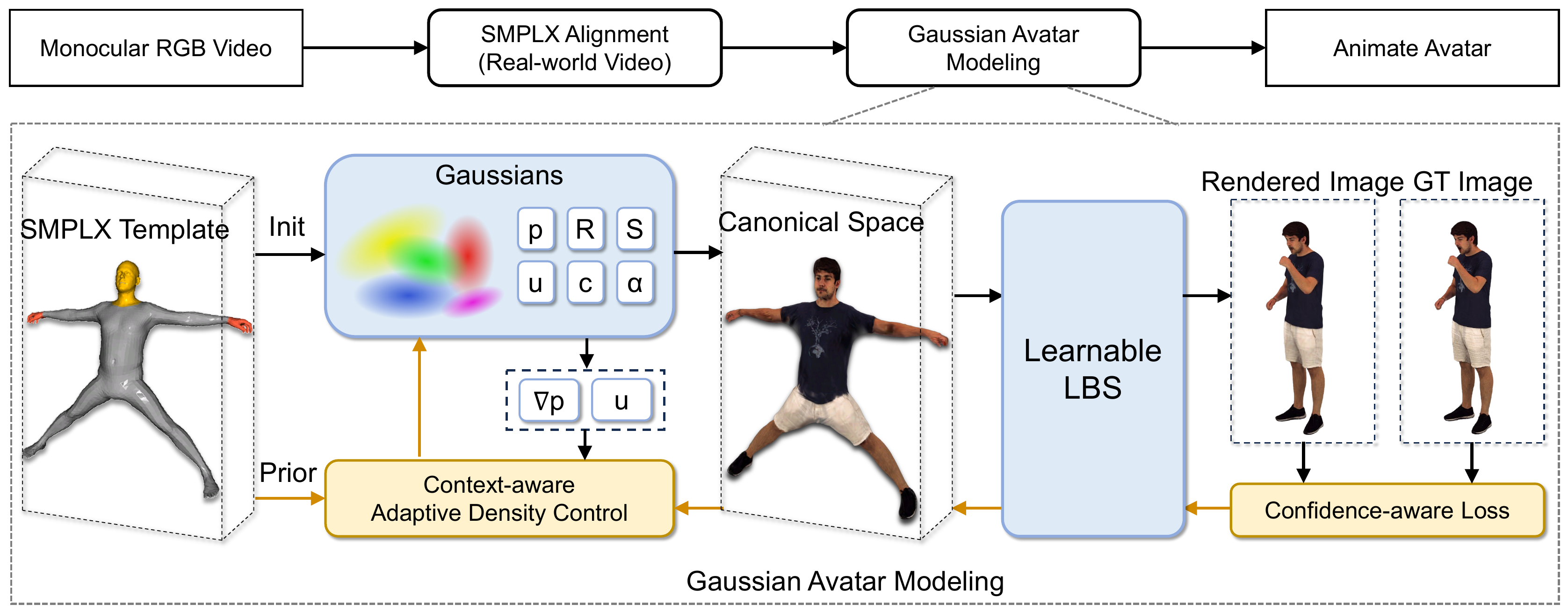}
    \caption{Overview of the proposed EVA framework. Given a real-world monocular RGB video, EVA first prepares well-aligned SMPL-X mesh via a plug-and-play module. Then EVA utilizes 3D Gaussians Splatting to perform avatar modeling, with the prior incorporated from the SMPL-X model. To improve the optimization, we propose context-aware adaptive density control and confidence-aware loss to improve the expressiveness of the avatar.}
    \label{fig:overview}
\end{figure}

\subsection{Context-aware Adaptive Density Control}
The core of optimizing Gaussian Splatting is adaptive density control, which generally contains two operations, \emph{i.e.,} densification~(split and clone) and pruning.
The original strategy~\cite{kerbl20233d} selects a fixed constant as the densification criteria.
However, utilizing the fixed threshold does not leverage context information, leading to sub-optimal Gaussian representations.

To this end, we propose a context-aware adaptive density control that leverages part attribute and history gradient information.
Each Gaussian inherently possesses attributes associated with different body parts. 
These body parts vary in spatial size and characteristics. 
For instance, compared to the body, the hand is much smaller in size, exhibiting fine-grained textures and a high degree of freedom. 
It also leads to inherent differences in the Gaussian positional gradient changes across different body parts during the Gaussian optimization process. 
Furthermore, a continuous increase in the Gaussian positional gradient indicates that Gaussians need to be densified. 

Specifically, we first initialize the 3D Gaussians with the vertices of the SMPL-X model.
Since the SMPL-X human model has the predefined topology, this initialization will give each Gaussian its attribute information $U$ on which part it belongs, \emph{e.g.} body, hand or face.
After that, considering attributes and gradient history information, the densification threshold for the $i$-th Gaussian in a certain attribute $U$ is formulated as follows:
\begin{equation}
\begin{aligned}
   \epsilon_{i} = e + \frac{\lambda_{t}}{R} (\sum_{k=t-R}^{t}\nabla_{i,k} - \sum_{k=t-2R}^{t-R}\nabla_{i,k}),
\label{eq:adc}
\end{aligned}
\end{equation}
where $e$ is a constant and $R$ represents the densification interval, and $\nabla_k$ denotes the position gradient of  $i$-th Gaussian on at the timestamp $k$.
Note that $e$ and ${\lambda_t}$ have different values for different attributes.
For the pruning strategy, we further append the human model prior via removing the points that are far away from the SMPL-X template vertices.

\subsection{Objective Functions}
The whole framework is optimized under the objective functions as follows,
\begin{equation}
\begin{aligned}
   \mathcal L= \mathcal L_{c} + \lambda_{m}\mathcal L_{m} + \lambda_{s}\mathcal L_{SSIM} + \lambda_{l}\mathcal L_{LPIPS},
\label{eq:loss_term}
\end{aligned}
\end{equation}
where $\lambda_{m}$, $\lambda_{s}$, and $\lambda_{l}$ are loss weighting terms.
The mask loss $\mathcal L_{m}$ calculates the consistency between accumulated volume density and mask pseudo ground truth.
The SSIM loss $\mathcal L_{SSIM}$ is adopted to improve the structural similarity between rendered image and ground truth image.
The LPIPS loss $\mathcal L_{LPIPS}$ focuses on the perceptual quality of the rendered image.

\noindent \textbf{Confidence-aware Loss $\mathcal L_{c}$.}
Given a training video, it is inevitable that it will contain some disturbance factors (\emph{e.g.}, misalignment, motion blur), which make it non-trivial to optimize the avatar in the canonical space via observations across different frames.
Therefore, we are motivated to propose a feedback module to inform the framework which pixel should have high confidence, and thus, should be taken into consideration with higher weight.
This module takes the rendered image and rendered depth as the input, and predicts a score for each pixel which represents the confidence as follows:
\begin{equation}
\begin{aligned}
   C = \mu + exp(E(I_r, D_r)),
\label{eq:conf}
\end{aligned}
\end{equation}
where $\mu$ is a constant.
$I_r$ and $D_r$ are the rendered image and rendered depth, respectively.
Then the confidence serves as an adaptive weighting factor on the per-pixel consistency. Eventually, we formulate our confidence-aware loss as follows:
\begin{equation}
\begin{aligned}
   \mathcal L_{c} = C \odot |I_r - I|_1.
\label{eq:loss_conf}
\end{aligned}
\end{equation}

\section{Experiments}
\label{sec_experiment}
In this section, we first introduce the experimental setup, including datasets, implementation details and evaluation metrics.
Then we make comparisons with baseline methods both quantitatively and qualitatively.
Finally, we perform ablation studies on the most important components of our framework.

\subsection{Experimental Setup}
\label{sec_setup}
\noindent \textbf{Datasets.}
The experiments are conducted on two datasets, \emph{i.e.,} \textbf{XHumans} and our collected \textbf{UPB} dataset.
\textbf{XHumans}~\cite{shen2023xavatar} is captured in the controlled environment.
It provides images with a resolution of 1200 $\times$ 800, along with well-aligned SMPL-X annotations.
There are 6 identities (3 male and 3 female) for evaluation.
\textbf{UPB} is collected from sign language videos in the web, which usually contain complicated hand gestures.
The resolution of the videos is 1920 $\times$ 1080 and do not contain any pose/SMPL-X annotations.
UPB contains 4 identities (2 male and 2 female).

\noindent \textbf{Implementation details.}
Our framework is implemented on PyTorch and all experiments are performed on NVIDIA A5000.
The hyperparameter $\lambda_m$, $\lambda_s$ and $\lambda_l$ are set to 0.1, 0.01 and 0.04, respectively.
We use SGHM~\cite{chen2022robust} to extract the human mask.
The Gaussian optimization lasts 2,000 iterations with the densification performed between 400 and 1,000 iterations.
For other parameters, we follow the original settings~\cite{kerbl20233d}.
We present more details in the Technical Appendices.

\noindent \textbf{Evaluation metrics.}
We evaluate the avatar quality through the rendered views following previous works~\cite{lei2023gart, GauHuman}.
We adopt the widely-used Peak Signal-to-Noise Ratio (PSNR), Structural Similarity Index Measure (SSIM)~\cite{wang2004image}, and Learned Perceptual Similarity (LPIPS)~\cite{zhang2018unreasonable} to evaluate the rendered images.
For a more fine-grained evaluation, we report the above metrics on the full body, on the hands and on the face regions, respectively.

\begin{table}[t]
    \footnotesize
    \centering
    \setlength{\tabcolsep}{0.65pt}
    \caption{Comparison with two expressive avatar baselines, \emph{i.e.}, GART + SMPLX and GauHuman + SMPLX, on the XHumans and UPB dataset. N-GS denotes the number of Gaussians. $\uparrow$ and $\downarrow$ represent the higher the better, and the lower the better, respectively.}
    \vspace{-0.5em}
    \begin{tabular}{l c ccc ccc ccc}
    \toprule
    \multirow{2}{*}{\textbf{Method}}  & \multirow{2}{*}{\textbf{N-GS}}  & \multicolumn{3}{c}{\textbf{Full}\quad\quad} & \multicolumn{3}{c}{\textbf{Hand}\quad\quad} & \multicolumn{3}{c}{\textbf{Face}\quad\quad} \\ 
    \cmidrule(lr){3-5} \cmidrule(lr){6-8} \cmidrule(lr){9-11}
    & & {\bf PSNR}$\uparrow$ & {\bf SSIM}$\uparrow$ & {\bf LPIPS}$\downarrow$ & {\bf PSNR}$\uparrow$ & {\bf SSIM}$\uparrow$ & {\bf LPIPS}$\downarrow$ & {\bf PSNR}$\uparrow$ & {\bf SSIM}$\uparrow$ & {\bf LPIPS}$\downarrow$ \\ \midrule
    \multicolumn{3}{l}{\emph{Controlled setting: XHumans dataset}} \\
    {3DGS~\cite{kerbl20233d} + SMPLX}  & 19,458 & 28.88 & 0.9609 & 44.93 & 25.28 & 0.9189 & 91.37 & 25.91 & 0.9087 & 101.04 \\
    {GART~\cite{lei2023gart} + SMPLX}  & 89,571 & 27.73 & 0.9553 & 50.32 & 25.42 & 0.9151 & 99.53 & 25.86 & 0.9013 & 105.06 \\
    {GauHuman~\cite{GauHuman} + SMPLX} & 17,134 & 29.16 & 0.9623 & 41.16 & 25.69 & 0.9225 & 88.16 & 26.27 & 0.9124 & 93.35 \\
    {EVA} & 19,993 & {\bf 29.67} & {\bf 0.9632} & {\bf 33.05} & {\bf 26.27} & {\bf 0.9279} & {\bf 72.95} & {\bf 26.56} & {\bf 0.9157} & {\bf 72.30} \\
    \midrule
    \multicolumn{3}{l}{\emph{Real-world setting: UPB dataset}} \\
    {3DGS~\cite{kerbl20233d} + SMPLX}  & 21,008 & 25.31 & 0.9469 & 90.80 & 24.89 & 0.9425 & 66.19 & 24.57 & 0.9072 & 136.53 \\
    {GART~\cite{lei2023gart} + SMPLX}  & 90,676 & 26.20 & 0.9511 & 78.90 & 25.25 & 0.9411 & 61.44 & 26.62 & 0.9253 & 93.28 \\
    {GauHuman~\cite{GauHuman} + SMPLX} & 12,372 & 25.17 & 0.9455 & 84.87 & 24.67 & 0.9418 & 67.61 & 24.33 & 0.9035 & 113.13 \\
    {EVA} & 20,829 & {\bf 26.78} & {\bf 0.9519} & {\bf 65.07} & {\bf 27.00} & {\bf 0.9524} & {\bf 45.90} & {\bf 26.85} & {\bf 0.9298} & {\bf 65.90} \\    
    \bottomrule
    \end{tabular}
    \label{tab:comparison}
    \vspace{-1.5em}
\end{table}

\subsection{Comparison with baselines}
The baselines we use for comparison are adopted from related methods with a few modifications.
3DGS + SMPL-X is modified from 3DGS~\cite{kerbl20233d}.
We add our articulated human modeling mechanism to make it fit the task requirement.
GART and GauHuman are originally designed to model body-level avatars.
They are animated via the SMPL parametric model, which lacks the capability of modeling hands and facial expressions.
Therefore, we replace the driven signal with the SMPL-X model and denote them as GART + SMPL-X and GauHuman + SMPL-X, respectively.
Among these baselines, GauHuman is mostly related to our approach.
Our differences are mainly reflected in the adaptive density control strategy, objective functions and the SMPL-X alignment for the real-world videos, which are also consistent with our main technical contributions.

\begin{figure}[t!]
    \centering
    \includegraphics[width=\linewidth]{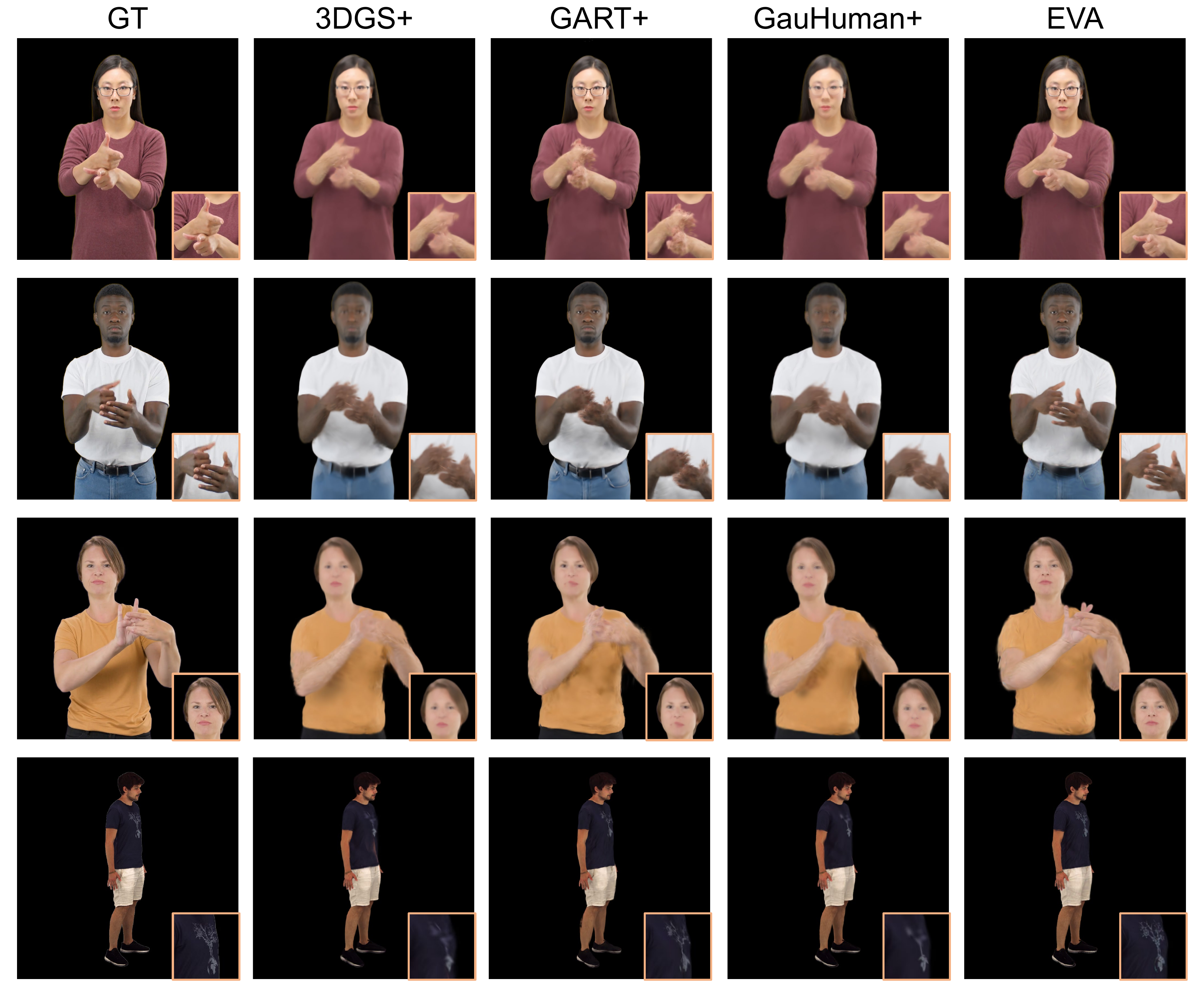}
    \caption{Qualitative comparison with baselines, including 3DGS~\cite{kerbl20233d} + SMPLX, GART~\cite{lei2023gart} + SMPLX, and GauHuman~\cite{GauHuman} + SMPLX on XHumans and UPB datasets. Our EVA model exhibits the best visual quality. See the zooming box for comparison of the fine-grained details.}
    \label{fig:vis}
    \vspace{-1.5em}
\end{figure}

As shown in Table~\ref{tab:comparison}, we conduct experiments on the XHumans and UPB datasets to evaluate the effectiveness of our method.
The former is captured in the controlled setting with accurate SMPL-X annotations, while the latter is oriented at the accessible video in daily life without pose information.
Since accurate SMPL-X annotation is hard to get in real-world scenarios, UPB is a more challenging benchmark.
We observe that our method achieves state-of-the-art performance on these two datasets.
Since the XHumans dataset contains SMPL-X annotation, we do not apply our designed alignment module.
Our method achieves 19.7\%, 17.3\% and 22.5\% relative LPIPS gain on the full, hand and face regions, respectively.
On the real-world UPB dataset, our performance gain is much larger, achieving over 25\% relative LPIPS gain on the hand region.
It indicates that EVA well adapts to the challenges from real-world videos, unlike previous work.
The qualitative comparison in Figure~\ref{fig:vis} also validates the effectiveness of our method.

\subsection{Ablation Studies}
We perform ablation experiments to highlight the important components of EVA, \emph{i.e.}, context-aware adaptive density control, confidence-aware loss and in-the-wild SMPL-X alignment module.

\noindent \textbf{Effectiveness of context-aware adaptive density control~(CADC) and confidence-aware loss~(CL).}
These two components are designed for better optimization of \emph{Gaussian Avatar Modeling}.
As shown in Table~\ref{tab:ablation_1}, ``w/o CADC'' means that we replace it with the original density control~\cite{kerbl20233d}, while ``w/o CL'' denotes removing the confidence weighting term in the RGB loss calculation.
We observe that these two components both improve performance.
More specifically, CADC improves performance for the metrics of all regions.

\begin{figure}[t!]
    \centering
    \includegraphics[width=\linewidth]{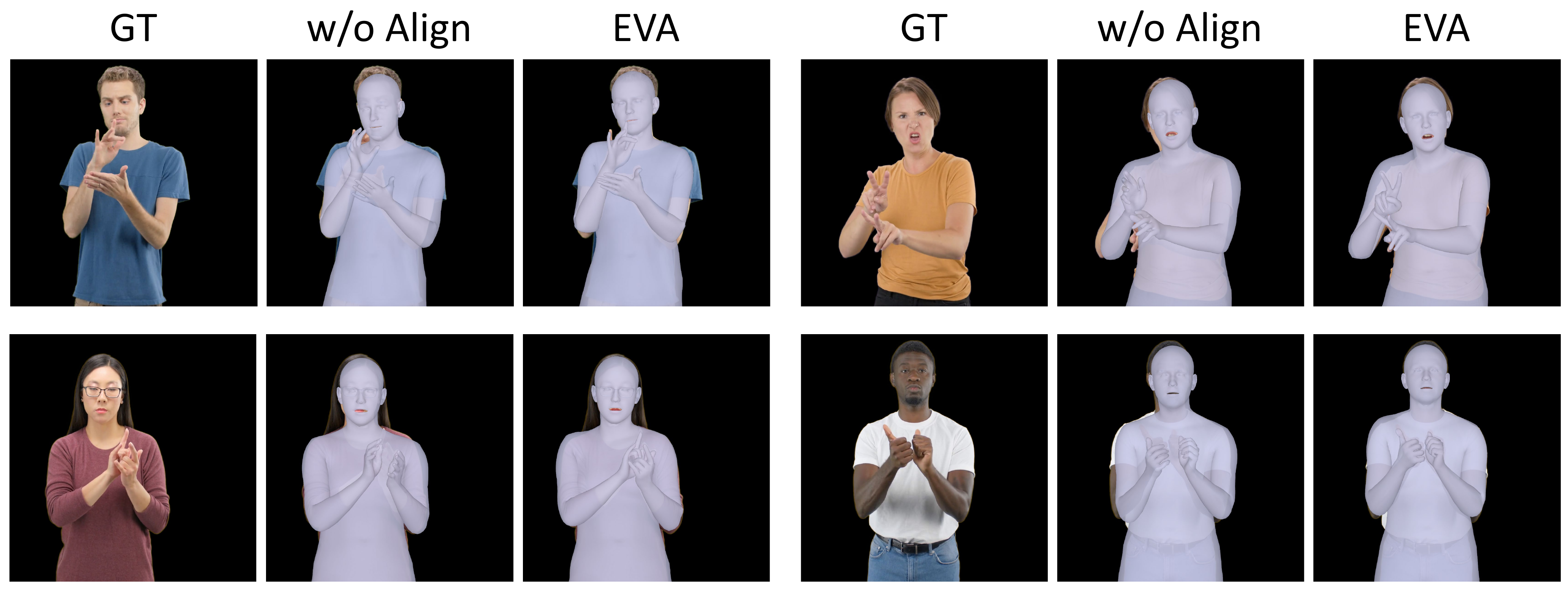}
    \caption{Demonstration of the effectiveness of our SMPL-X alignment module. We can produce a SMPL-X mesh that aligns well with the RGB frame, especially for the fine-grained hand regions.}
    \label{fig:smplx_align}
\end{figure}

\noindent \textbf{Effectiveness of in-the-wild SMPL-X alignment.}
Our SMPL-X alignment module is utilized for the real-world video without accurate SMPL-X mesh and we conduct this ablation study on UPB dataset.
As shown in Table~\ref{tab:ablation_2}, ``w/o Align'' means that we directly utilize the SMPL-X mesh extracted from current SOTA estimation method~\cite{cai2024smpler}.
Notably, our proposed alignment module brings notable performance gains on all metrics of all regions.
It also demonstrates the importance of the SMPL-X pose quality for final human avatar modeling.
Without accurate SMPL-X alignment, the following Gaussian model will need to deal with inconsistent textures for a specific region across frames, leading to degraded avatar quality. 
Furthermore, we visualize the SMPL-X alignments in Figure~\ref{fig:smplx_align}, 
where we observe visible improvements specifically for the hands.

\begin{table}[t]
    \footnotesize
    \centering
    \setlength{\tabcolsep}{5pt}
    \caption{Ablation study on context-aware adaptive density control~(CADC) and confidence-aware loss~(CL) on XHumans. $\uparrow$ and $\downarrow$ represent ``higher the better'', and ``lower the better'', respectively.}
    \begin{tabular}{l ccc ccc ccc}
    \toprule
    \multirow{2}{*}{\textbf{Method}}  & \multicolumn{3}{c}{\textbf{Full}\quad\quad} & \multicolumn{3}{c}{\textbf{Hand}\quad\quad} & \multicolumn{3}{c}{\textbf{Face}\quad\quad} \\ 
    \cmidrule(lr){2-4} \cmidrule(lr){5-7} \cmidrule(lr){8-10}
    & {\bf PSNR}$\uparrow$ & {\bf SSIM}$\uparrow$ & {\bf LPIPS}$\downarrow$ & {\bf PSNR}$\uparrow$ & {\bf SSIM}$\uparrow$ & {\bf LPIPS}$\downarrow$ & {\bf PSNR}$\uparrow$ & {\bf SSIM}$\uparrow$ & {\bf LPIPS}$\downarrow$ \\ \midrule
    {w/o CADC} & 28.92 & 0.9611 & 35.14 & 25.23 & 0.9175 & 84.00 & 26.08 & 0.9080 & 80.76 \\
    {w/o CL} & 29.63 & {\bf 0.9634} & 34.08 & 26.27 & 0.9279 & 74.02 & {\bf 26.58} & {\bf 0.9166} & 73.19 \\
    {EVA}& {\bf 29.66} & {0.9632} & {\bf 33.05} & {\bf 26.27} & {\bf 0.9279} & {\bf 72.95} & {26.56} & {0.9156} & {\bf 72.30} \\
    \bottomrule
    \end{tabular}
    \label{tab:ablation_1}
\end{table}

\begin{table}[t!]
    \footnotesize
    \centering
    \setlength{\tabcolsep}{5pt}
    \caption{Ablation study of SMPL-X alignment on the UPB benchmark. $\uparrow$ and $\downarrow$ represent ``higher the better'', and ``lower the better'', respectively.}
    \begin{tabular}{l ccc ccc ccc}
    \toprule
    \multirow{2}{*}{\textbf{Method}}  & \multicolumn{3}{c}{\textbf{Full}\quad\quad} & \multicolumn{3}{c}{\textbf{Hand}\quad\quad} & \multicolumn{3}{c}{\textbf{Face}\quad\quad} \\ 
    \cmidrule(lr){2-4} \cmidrule(lr){5-7} \cmidrule(lr){8-10}
    & {\bf PSNR}$\uparrow$ & {\bf SSIM}$\uparrow$ & {\bf LPIPS}$\downarrow$ & {\bf PSNR}$\uparrow$ & {\bf SSIM}$\uparrow$ & {\bf LPIPS}$\downarrow$ & {\bf PSNR}$\uparrow$ & {\bf SSIM}$\uparrow$ & {\bf LPIPS}$\downarrow$ \\ \midrule
    {w/o Align} & 25.02 & 0.9435 & 73.82 & 24.64 & 0.9396 & 63.64 & 24.27 & 0.9009 & 93.56 \\
    {EVA} & {\bf 26.72} & {\bf 0.9519} & {\bf 65.37} & {\bf 26.90} & {\bf 0.9523} & {\bf 46.11} & {\bf 26.75} & {\bf 0.9298} & {\bf 66.41} \\
    \bottomrule
    \end{tabular}
    \label{tab:ablation_2}
\end{table}

\subsection{Limitations and Broader Impact}
\label{sec_limitation}
In this work, we are dedicated to the expressiveness of the human avatar and designing a pipeline that could take the raw Web video as the input.
Besides expressiveness, some other factors are challenging yet worth investigating to further improve the quality of the built avatar, \emph{e.g.} non-rigid elements like cloth and hair.
It is also worth extending avatar reconstruction with more challenging input sources, \emph{e.g.} occlusion or few-shot scenarios.

Our research could create a 3D digital avatar with low cost (only need monocular RGB video), which benefits multiple application scenarios, \emph{e.g.} AR/VR, movie production, video games, \emph{etc.}
Its outperforming capability of expressiveness better conveys the semantic meaning, thereby optimizing the user experience.
On the other side, our framework could generate non-existent images.
It may be misused in fake content creation and lead to privacy leakage, which is opposite to our natural intent.
It can be tackled via various regulations and technical measures (\emph{e.g.} watermarking).

\section{Conclusion}
In this work, we present EVA, a drivable expressive human avatar learned from real-world monocular RGB video.
EVA is built on 3D Gaussian, in coordination with human prior introduced by SMPL-X.
To deal with the challenges brought by expressiveness, we first utilize a plug-and-play module to solve the misalignment between pose and RGB frame.
During the optimization of 3D Gaussian, we propose context-aware adaptive density control, which leverages attribute and historical gradient information to accommodate the varied granularity across body parts.
A feedback mechanism is jointly designed to further guide the learning.
Extensive experiments on two benchmarks demonstrate our method outperforms baselines both quantitatively and qualitatively, especially on the fine-grained hand and facial details.

{\small
\bibliographystyle{ieee_fullname}
\bibliography{egbib}
}

\newpage
\section*{Technical Appendices}
\label{sec:supp}
This technical appendices provides more details which are not included in the main paper due to space limitations.
For more visualization results, please refer to the attached video file named ``Supp.mp4''.

\textbf{Training/Testing Split.}
For each identity in XHumans dataset, one video is selected as the training split, where the other videos are marked as testing.
During training, we utilize all of the frames (150 frames).
We sample 20 frames for each testing video with the sampling rate of 5.
For UPB dataset, we uniformly sample the frames with the interval as 1 to split the training and testing frames.
The number of training and testing frames are both 140.

\textbf{Additional Implementation Details.}
$\mu$ is set to 1.
$\lambda_t$ is set as -9.0, -4.5 and -6.3 for body, hand, and face parts, respectively.
We set $e$ for body, hand, and face parts as 2e-4, 1e-4 and 1.4e-4, respectively.
For SMPL-X alignment, we utilize the L-BFGS optimizer with the Wolfe line search.
The optimization contains three stages with different loss weighting factors.
The first stage is designed to initialize the human body embedding to the initial estimation from the method~\cite{cai2024smpler}.
Then the second stage mainly aims to get better spatial hand relationship leveraging the prediction from the method~\cite{pavlakos2023reconstructing}.
The third stage performs fine-tuning with more emphasize on the fine-grained hand and face regions.

\textbf{Comparison Baselines.}
The baselines we use for comparison are adopted from related methods with a few modifications.
3DGS + SMPL-X is modified from 3DGS~\cite{kerbl20233d}.
Since the original 3DGS does not have the capability of animation, we add the same articulated human modeling in Section~\ref{sec:prelim} as ours.
For fair comparison, it also utilizes the same optimization schedule as ours, which lasts 2,000 iterations with the densification performed between 400 and 1,000 iterations. 
GART and GauHuman are originally designed to model body-level avatars.
They are animated via the SMPL parametric model, which lacks the capability of modeling hands and facial expressions.
Therefore, we replace the driven signal with the SMPL-X model and denote them as GART + SMPL-X and GauHuman + SMPL-X, respectively.
We do not modify the optimization schedules of these two methods.

\end{document}